\documentclass{article}

\usepackage{microtype}
\usepackage{graphicx}
\usepackage{subcaption}
\usepackage{booktabs} 
\usepackage{tikz}
\usepackage{xcolor}
\usepackage{tabularx}
\usepackage[table]{xcolor}
\usetikzlibrary{shapes.geometric, arrows.meta, positioning, fit, backgrounds, calc, shadows.blur}

\usepackage[switch]{lineno}
\usepackage{bm}

\usepackage{hyperref}

 \usepackage{siunitx} 
 \usepackage{xcolor,colortbl}
\usepackage[preprint]{icml2026}

\usepackage{amsmath}
\usepackage{amssymb}
\usepackage{mathtools}
\usepackage{amsthm}

\usepackage[capitalize,noabbrev]{cleveref}

\theoremstyle{plain}

\theoremstyle{definition}

\theoremstyle{remark}

\usepackage[textsize=tiny]{todonotes}
\definecolor{bestbg}{RGB}{220,245,230}      
\definecolor{secondbg}{RGB}{235,245,255}    
\definecolor{poorbg}{RGB}{255,238,238}      
\icmltitlerunning{SCALAR: Quantifying Structural Hallucination, Consistency, and Reasoning in Materials Foundation Models}

\begin{document}

\twocolumn[
  \icmltitle{SCALAR: Quantifying Structural Hallucination, Consistency, and\\ Reasoning Gaps in Materials Foundation Models}




  \begin{icmlauthorlist}
  \icmlauthor{Can Polat}{tamu}
  \icmlauthor{Erchin Serpedin}{tamu}
  \icmlauthor{Mustafa Kurban}{ankara,tamuq}
  \icmlauthor{Hasan Kurban}{hbku}
\end{icmlauthorlist}

\icmlaffiliation{tamu}{Texas A\&M University, College Station, TX, USA}
\icmlaffiliation{ankara}{Ankara University, Ankara, Turkey}
\icmlaffiliation{tamuq}{Texas A\&M University at Qatar, Doha, Qatar}
\icmlaffiliation{hbku}{Hamad Bin Khalifa University, Doha, Qatar}

\icmlcorrespondingauthor{Mustafa Kurban}{kurbanm@ankara.edu.tr}
\icmlcorrespondingauthor{Hasan Kurban}{hkurban@hbku.edu.qa}

  \icmlkeywords{dft, materials science, crystals, llms, foundational models, structural hallucination, reasoning gaps}

  \vskip 0.3in
]



\printAffiliationsAndNotice{}  

\begin{abstract}
Large language models are increasingly applied to materials science reasoning, yet their behavior under physically structured distribution shifts remains poorly understood. We introduce SCALAR (Structural Consistency And Logic Across Regimes), a benchmark for evaluating geometric scale generalization and its connection to structural hallucination, consistency, and reasoning in materials foundation models. Given canonical crystal representations, models must reason about derived nanoparticle structures obtained through supercell expansion and geometric truncation across length scales spanning a few atoms to over 18,000 atoms, totaling $\approx$100,000 structures from DFT-validated unit cells. SCALAR defines three tasks. (i) CIF to property prediction. (ii) A Chain-of-Thought variant with explicit physics-grounded reasoning. (iii) Inverse retrieval identifying crystals from candidates given target properties. Outputs are evaluated via structured metrics capturing numeric error, hallucination, cross-prompt consistency, monotonic reasoning, output validity, and retrieval regret. Experiments across diverse foundation models reveal large, model-dependent shifts under explicit reasoning, often reducing hallucination and error, but frequently destabilizing consistency or validity. These results demonstrate that geometric scale generalization cannot be inferred from accuracy alone. Supplementary materials are available at \url{https://github.com/KurbanIntelligenceLab/SCALAR}.

\end{abstract}

\section{Introduction}
Materials modelling has historically been divided between two complementary but distinct regimes. The first is the regime of perfect crystals, where structures are represented by a unit cell defined through lattice constants, atomic positions, and space group symmetry \citep{tarantino2017structural}. This abstraction, codified in CIFs, captures the periodic order of an infinite solid and underpins the foundations of solid-state physics and computational chemistry \citep{kittel2018introduction, ashcroft1976solid}. The second regime is that of nanoparticles, finite clusters of atoms that break translational invariance. In nanoparticles, surfaces, edges, and under-coordinated sites dominate, leading to reconstructions, distortions, and quantum confinement effects that strongly alter material properties \citep{pizzagalli2001structure, bera2010quantum}. Real materials often bridge these two representations, and understanding the transition between them is critical for predicting catalytic activity, optical response, stability, and electronic behavior \citep{vergara2017microed}.

\begin{figure}[ht]
\centering
\resizebox{1\columnwidth}{!}{%
\begin{tikzpicture}[
  font=\sffamily\small,
  >=Stealth,
  main_box/.style={
    draw=#1!70!black, fill=#1!8, line width=0.8pt,
    rounded corners=3pt, inner sep=6pt,
    minimum height=1.4cm,
    minimum width=4cm,
    text width=3.5cm, align=center,
    drop shadow={shadow xshift=0.5pt, shadow yshift=-0.5pt, opacity=0.15}
  },
  phase_label/.style={
    fill=#1!25, draw=#1!60!black, line width=0.6pt,
    font=\bfseries\small, inner sep=5pt, rounded corners=2pt,
    minimum width=4cm
  },
  step_label/.style={
    font=\sffamily\scriptsize\itshape, text=black!70
  },
  annotation_box/.style={
    draw=#1!50, fill=#1!5, rounded corners=2pt, 
    font=\scriptsize, inner sep=4pt, text width=2.2cm, align=center
  },
  flow_arrow/.style={
    ->, draw=black!60, line width=1.2pt, shorten >=3pt, shorten <=3pt
  }
]

\definecolor{phase1}{RGB}{16,124,65}    
\definecolor{phase2}{RGB}{0,90,156}     
\definecolor{phase3}{RGB}{148,70,140}   

\def\colsep{5cm}
\def\rowsep{1.5cm}

\node[phase_label=phase1] (P1) at (0, 0) {\textsc{Phase I}};
\node[font=\normalsize, below=1pt of P1] (P1sub) {Nanoparticle Construction};

\node[main_box=phase1, below=0.8cm of P1sub] (UC) {
  \begin{tabular}{c}
    \textbf{Unit Cell} $\mathcal{U}$ \\[2pt]
    {\footnotesize $\Lambda=(a,b,c,\alpha,\beta,\gamma)$}
  \end{tabular}
};

\node[main_box=phase1, below=\rowsep of UC] (SC) {
  \begin{tabular}{c}
    \textbf{Supercell} $\mathcal{T}$ \\[2pt]
    {\footnotesize $L_i = 20\,a_i \ge 2R_{\max}+\Delta$}
  \end{tabular}
};

\node[main_box=phase1, below=\rowsep of SC] (NP) {
  \begin{tabular}{c}
    \textbf{Nanoparticle} $\mathcal{C}_R$ \\[2pt]
    {\footnotesize $\|\mathbf{x}_i-\mathbf{x}_0\|\le R$}
  \end{tabular}
};

\draw[flow_arrow] (UC) -- (SC) node[midway, right, step_label, xshift=2pt] {replicate};
\draw[flow_arrow] (SC) -- (NP) node[midway, right, step_label, xshift=2pt] {carve};

\node[phase_label=phase2] (P2) at (\colsep, 0) {\textsc{Phase II}};
\node[font=\normalsize, below=1pt of P2] (P2sub) {$\mathrm{SO}(3)$ Rotation Sampling};

\node[main_box=phase2, below=0.8cm of P2sub] (QS) {
  \begin{tabular}{c}
    \textbf{Quaternion} $\mathbf{u}$ \\[2pt]
    {\footnotesize $\mathbf{u}\in\mathbb{H},\;\|\mathbf{u}\|=1$}
  \end{tabular}
};

\node[main_box=phase2, below=\rowsep of QS] (GC) {
  \begin{tabular}{c}
    \textbf{Geodesic Spacing} \\[2pt]
    {\footnotesize $\Delta(\mathbf{u}_i,\mathbf{u}_j)\ge\vartheta$}
  \end{tabular}
};

\node[main_box=phase2, below=\rowsep of GC] (ER) {
  \begin{tabular}{c}
    \textbf{Effective Rotation} \\[2pt]
    {\footnotesize $\mathbf{u}_{\mathrm{eff}}=\mathbf{u}_{\mathrm{off}}\mathbf{u}$}
  \end{tabular}
};

\draw[flow_arrow] (QS) -- (GC) node[midway, right, step_label, xshift=2pt] {enforce};
\draw[flow_arrow] (GC) -- (ER) node[midway, right, step_label, xshift=2pt] {offset};

\node[phase_label=phase3] (P3) at (2*\colsep, 0) {\textsc{Phase III}};
\node[font=\normalsize, below=1pt of P3] (P3sub) {Split-Aware Partitioning};

\node[main_box=phase3, below=0.8cm of P3sub] (TR) {
  \begin{tabular}{c}
    \textbf{Train} \\[2pt]
    {\footnotesize $\vartheta_{\mathrm{tr}}=22^\circ$, $K_{\mathrm{tr}}=60$}
  \end{tabular}
};

\node[main_box=phase3, below=\rowsep of TR] (ID) {
  \begin{tabular}{c}
    \textbf{ID-Test} \\[2pt]
    {\footnotesize $\vartheta_{\mathrm{ID}}=18^\circ$, $\varepsilon_{\mathrm{ID}}=8^\circ$}
  \end{tabular}
};

\node[main_box=phase3, below=\rowsep of ID] (OD) {
  \begin{tabular}{c}
    \textbf{OOD-Test} \\[2pt]
    {\footnotesize $\vartheta_{\mathrm{OD}}\in\{16^\circ,28^\circ\}$}
  \end{tabular}
};

\draw[flow_arrow] (TR) -- (ID) node[midway, right, step_label, xshift=2pt] {exclude};
\draw[flow_arrow] (ID) -- (OD) node[midway, right, step_label, xshift=2pt] {exclude};

\begin{scope}[on background layer]
  \node[fill=phase1!4, draw=phase1!25, rounded corners=5pt, line width=0.5pt,
        fit=(P1) (P1sub) (UC) (SC) (NP),
        inner sep=10pt] (BG1) {};

  \node[fill=phase2!4, draw=phase2!25, rounded corners=5pt, line width=0.5pt,
        fit=(P2) (P2sub) (QS) (GC) (ER),
        inner sep=10pt] (BG2) {};

  \node[fill=phase3!4, draw=phase3!25, rounded corners=5pt, line width=0.5pt,
        fit=(P3) (P3sub) (TR) (ID) (OD),
        inner sep=10pt] (BG3) {};
\end{scope}

\end{tikzpicture}%
}
\caption{Systematic benchmark construction pipeline. \textbf{Phase~I} constructs finite nanoparticles $\mathcal{C}_R$ from DFT-relaxed unit cells via supercell replication and spherical carving at radii $R\in\{10,\ldots,30\}$\,\AA. \textbf{Phase~II} samples rigid rotations on $\mathrm{SO}(3)$ using unit quaternions with minimum geodesic spacing $\vartheta$, then applies split-specific offsets. \textbf{Phase~III} enforces split-aware exclusion: ID rotations maintain margin $\varepsilon_{\mathrm{ID}}$ from training, OOD rotations maintain margin $\varepsilon_{\mathrm{OD}}$ from both training and ID sets, with radius-based partitioning ($\mathcal{S}_{\mathrm{ID}}$, $\mathcal{S}_{\mathrm{OD}}$).}
\label{fig:methodology_pipeline}
\end{figure}
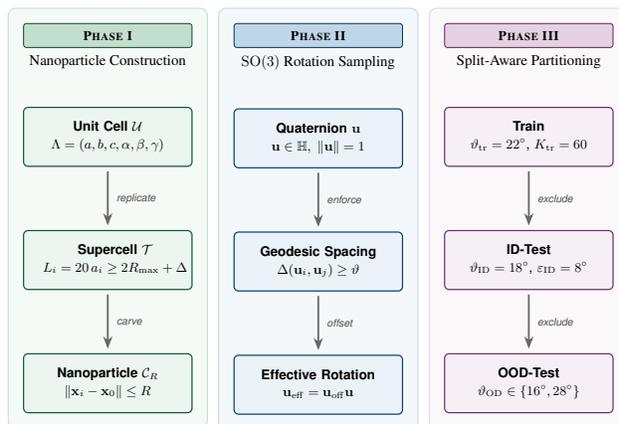

Despite advances in computational modelling and machine learning, bridging bulk and nanoscale regimes remains a major challenge \citep{li2023critical}. Models trained exclusively on bulk data often reproduce unit-cell properties accurately, but degrade when tasked with generating nanoscale structures or reconstructing bulk representations from finite, surface-distorted inputs \citep{gleason2024random}. More broadly, these failures reflect a limitation shared by many modern generative models, including large language models (LLMs), when applied to scientific domains \citep{ji2023survey}. Unlike natural language, materials structures are governed by latent global invariants—such as crystallographic symmetry, conservation constraints, and scale-dependent geometric relations—that must remain consistent across different views or resolutions of the same underlying object \citep{mirza2025framework}. When these invariants are violated, models can produce outputs that appear locally plausible yet are globally inconsistent or physically incorrect, a phenomenon commonly described as hallucination in scientific generation \citep{maynez2020faithfulness}. We argue that such errors constitute a principled form of hallucination: confident predictions that violate underlying physical invariants, particularly under distribution shifts induced by changes in structural scale.

Diagnosing structural hallucinations of this form requires datasets that provide paired representations of the same object across multiple scales, together with explicit notions of global invariants and controlled out-of-distribution splits. To address this need, we introduce a \emph{cross-scale evaluation framework} that explicitly couples bulk and finite representations, enabling systematic assessment of scale generalization and invariant preservation \citep{jablonka202314}. In this work, we instantiate this framework using a chemically diverse family of crystalline materials with well-defined bulk symmetry. Each material is represented by a compact unit cell and a large collection of systematically generated finite structures spanning radii from 10~\AA\ to 30~\AA, allowing controlled variation in size while holding the underlying crystallographic identity fixed.

\begin{figure*}[ht] 
\centering 
\includegraphics[width=\linewidth]{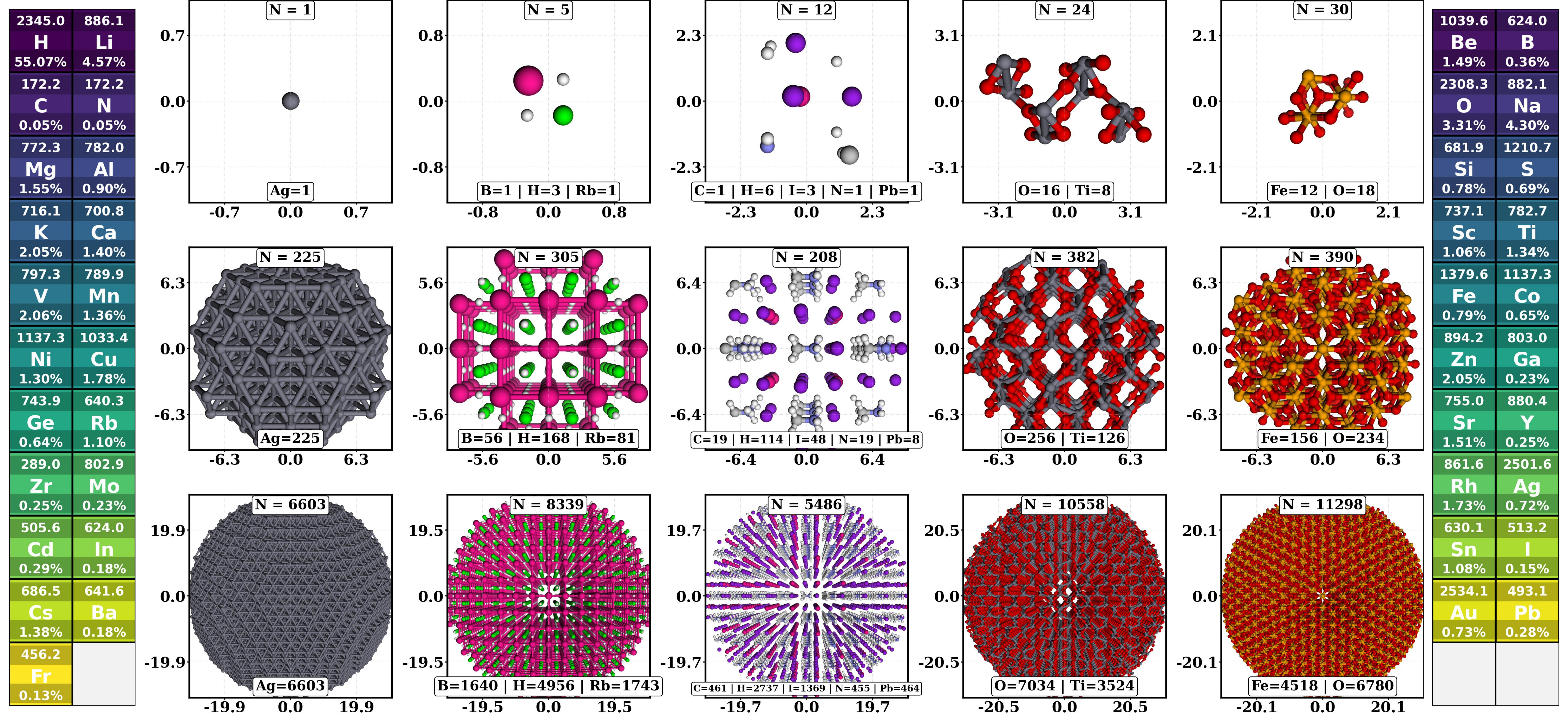} 
\caption{Columns (left to right) correspond to Ag, RbBH$_{3}$, CH$_{3}$NH$_{3}$PbI$_{3}$, TiO$_{2}$, Fe$_{2}$O$_{3}$. Rows show the reference unit cell (top), a nanoparticle carved at $R=10$\,\AA\ (middle), and a nanoparticle carved at $R=30$\,\AA\ (bottom). Each panel reports the total atom count ($N$) and elemental composition. Across the dataset, particle sizes span more than four orders of magnitude, from a minimum of $N=4$ atoms for the smallest particles to a maximum of $N=18,123$ atoms for the largest configurations. For the element columns, the value above each element is its percent of total atoms in the dataset, and the value below each element is the average atoms per structure containing that element. Atom colors follow the standard CPK scheme.}
\label{fig:elementsAndMaterials} 
\end{figure*}

\section{Related Work}\label{sec:relatedWork}

\subsection{Crystalline and Nanoscale Materials}

Crystalline solids are commonly described by a unit cell—the smallest repeating motif defined by lattice parameters, symmetry, and an atomic basis—which underlies crystallography, density functional theory, and most large materials databases \citep{anosova2024importance}. While this abstraction efficiently encodes infinite periodic order, many real materials exist at finite sizes \citep{baig2021nanomaterials}. At the nanoscale, nanoparticles break translational symmetry and exhibit strong surface effects, including under-coordination, reconstruction, and edge distortions, which substantially alter physical and chemical behavior \citep{zhang2023surface}. As a result, properties such as catalytic activity and optical or electronic response can differ markedly from bulk behavior due to surface sensitivity and quantum confinement \citep{ye2024strongly}. Relating bulk crystallographic descriptors to nanoscale structures—or recovering bulk invariants from finite particles—therefore remains a fundamental challenge, motivating cross-scale evaluations that probe consistency across size and representation.

\subsection{Evaluation and Cross-Scale Limitations}

Large-scale datasets have driven progress in machine learning for materials science by providing the structural and energetic data required for modern predictive and generative models. Resources such as the Materials Project \citep{jain2013commentary} and OQMD \citep{kirklin2015open} curate extensive collections of bulk crystal structures, while PubChemQC \citep{kim2025pubchem} provides large-scale quantum chemical data for molecular systems. Despite their breadth, these datasets are largely single-scale, typically representing materials either as infinite crystals or isolated molecules, and rarely include paired representations across systematically varied sizes. Efforts to address this gap, such as MultiSpectrumSet \citep{polat2025beyond}, remain limited in scope. Consequently, existing benchmarks are poorly suited for diagnosing whether models preserve global invariants—such as symmetry or structural identity—across resolutions, motivating cross-scale benchmarks that explicitly couple bulk and finite representations.

\subsection{From Geometry-Aware Generative Models to LLM-Based Foundation Models}

The availability of large structural datasets has enabled diverse machine learning approaches for materials modeling, with geometry-native models—particularly graph neural networks and equivariant architectures—achieving strong performance by explicitly encoding atomic structure and physical symmetries \citep{dimenet,klipfel2023equivariant, kurban2026multimodal}. More recently, generative models have expanded to hybrid approaches that combine continuous structure generation with language-based conditioning \citep{sriram2024flowllm}. In parallel, large language models and foundation models are increasingly used to generate molecular graphs, crystal descriptions, reaction pathways, and simulation inputs \citep{m2024augmenting, jablonka2024leveraging, ramos2025review}, including materials-specialized models based on domain-adapted pretraining \citep{mishra2024foundational, xie2024darwin}, multimodal integration of structural and property data \citep{moro2025multimodal}, and platforms for fine-tuning atomistic foundation models \citep{kong2025mattertune}. Operating primarily over symbolic or textual encodings, these models offer broad flexibility but make it difficult to enforce global physical invariants, leading to outputs that may appear locally plausible while violating global consistency, a phenomenon closely related to hallucination \citep{zheng2023chatgpt}. Existing evaluations largely focus on task accuracy and formatting correctness and rarely assess cross-scale consistency \citep{polat2025stress}, motivating evaluation frameworks that explicitly probe invariant preservation across representations alongside geometry-aware generative approaches.

\section{SCALAR}\label{sec:method}

\subsection{Nanoparticle construction}
\label{sec:nano_construction}

We curated the dataset with high hydrogen-storage potential and retrieved their \emph{DFT-relaxed} \citep{bickelhaupt2000kohn} lattice parameters from the \textit{experimental} results. From the corresponding CIFs we first built the primitive unit cell, then expanded to a supercell that accommodates the target radius. Finite nanoparticles were obtained by retaining atoms within a sphere of radius $R$ centered at $x_0$. No additional relaxation or surface/ligand modeling was applied, thereby isolating size/geometry effects while remaining consistent with vetted bulk parameters.

\noindent\textbf{Supercell size.}
For each composition, we constructed a $20\times20\times20$ supercell by replicating the primitive unit cell along $a,b,c$, yielding box lengths $L_i=20\,a_i$ ($i\in\{a,b,c\}$). This satisfies $L_i \ge 2R_{\max}+\Delta$ for the maximum carving radius $R_{\max}=30\,\text{\AA}$ with a safety margin $\Delta\approx 5$–$10\,\text{\AA}$, ensuring carved nanoparticles remain well-separated from periodic images.

\noindent\textbf{Spherical carving.}
A finite nanoparticle of target radius $R$ is constructed by retaining all atoms whose Cartesian positions fall within a sphere of radius $R$ centered at a chosen origin, with $R \in \{10,\dots,30\}$\,\AA\ and $\mathcal{C}_R=\{\,\mathbf{x}_i \in \mathbb{R}^3 \mid \|\mathbf{x}_i-\mathbf{x}_0\|\le R\,\}$. This procedure yields a controlled family of clusters per composition, spanning the transition from strongly surface-dominated (small $R$) to near-bulk behavior (large $R$).

\subsection{Rotation Sampling and Dataset Splits}

To mitigate orientation bias while enforcing split-dependent separation, each structure is augmented by rigid rotations sampled on $\mathrm{SO}(3)$ \citep{shoemake1985animating}. Rotations are represented by unit quaternions $\mathbf{u}\in\mathbb{H}$ with $\|\mathbf{u}\|=1$. The geodesic distance between two rotations $\mathbf{u}_i,\mathbf{u}_j$ is
\[
\Delta(\mathbf{u}_i,\mathbf{u}_j)
=2\arccos\!\left(\big|\mathbf{u}_i^\top \mathbf{u}_j\big|\right),
\]
where the absolute value accounts for the antipodal equivalence $\mathbf{u}\sim -\mathbf{u}$. A greedy sampler constructs a set $\mathcal{U}(\vartheta)$ such that
\[
\Delta(\mathbf{u}_i,\mathbf{u}_j)\ge \vartheta
\quad \forall\, i\neq j,
\]
with spacing parameter $\vartheta$ controlling the sampling density. To avoid uncontrolled growth in dataset size, the number of accepted rotations per structure is capped.

\paragraph{Reference training lattice.}
We first generate a deterministic reference lattice of rotations,
denoted $\mathcal{U}_{\mathrm{tr}}$, by sampling $K_{\mathrm{tr}}$ quaternions with minimum spacing $\vartheta_{\mathrm{tr}}=22^\circ$ (here $K_{\mathrm{tr}}=60$). For each training structure, a subset of size $k_{\mathrm{tr}}\le K_{\mathrm{tr}}$ is selected without replacement from $\mathcal{U}_{\mathrm{tr}}$ using a stable, structure-dependent seed.

\paragraph{Split-aware exclusion constraints.}
For evaluation splits, rotations are sampled greedily under two constraints: (i) an internal spacing condition and (ii) a margin from a reference set of previously realized rotations. Specifically, a candidate rotation $\mathbf{u}$ is accepted only if
$\Delta(\mathbf{u},\mathbf{v})\ge \varepsilon_{\mathrm{split}}$, $\forall\, \mathbf{v}\in \mathcal{U}_{\mathrm{ref}},$
which is equivalent to
\[
\big|\mathbf{u}^\top \mathbf{v}\big|
\le
\cos\!\Big(\tfrac{1}{2}\varepsilon_{\mathrm{split}}\Big).
\]
ID rotations are generated with spacing $\vartheta_{\mathrm{ID}}=18^\circ$ and exclusion margin $\varepsilon_{\mathrm{ID}}=8^\circ$ from the training reference set $\mathcal{U}_{\mathrm{tr}}$. OOD rotations are generated using two regimes: a dense regime with spacing $\vartheta_{\mathrm{OD,d}}=16^\circ$ and a sparse regime with spacing $\vartheta_{\mathrm{OD,s}}=28^\circ$. Both OOD regimes enforce an exclusion margin $\varepsilon_{\mathrm{OD}}=8^\circ$ from the union $\mathcal{U}_{\mathrm{tr}}\cup\mathcal{U}_{\mathrm{ID}}$, where $\mathcal{U}_{\mathrm{ID}}$ denotes the set of ID rotations actually realized after symmetry-induced deduplication.

\paragraph{Deterministic rotation offsets.}
To further decorrelate the orientation distributions across splits, we apply a fixed left-multiplication to all sampled rotations in each non-training split:
\[
\mathbf{u}_{\mathrm{eff}} = \mathbf{u}_{\mathrm{off},\mathrm{split}} \,\mathbf{u},
\]
where $\mathbf{u}_{\mathrm{off},\mathrm{ID}}$ and $\mathbf{u}_{\mathrm{off},\mathrm{OD}}$ are fixed quaternions corresponding to Euler offsets  $(12^\circ,16^\circ,24^\circ)$ and $(30^\circ,50^\circ,70^\circ)$, respectively. These offsets do not by themselves guarantee separation; strict disjointness is enforced by the exclusion margins $\varepsilon_{\mathrm{split}}$.

\paragraph{Radius-based partitioning and budgets.}
Structures are partitioned by radius $R$ into
$\mathcal{S}_{\mathrm{ID}}=\{13,15,17,20,24,27\},\;
 \mathcal{S}_{\mathrm{OOD}}=\{10,11,29,30\}$.
with all remaining radii assigned to training. For each dataset, we fix a target total number of generated structures (including originals) and allocate this target across splits using predetermined fractions. Per-structure augmentation counts are derived from the number of base structures in each split and are capped by split-specific maxima. For OOD, the augmentation budget is further divided between dense and sparse regimes. Finally, to account for point-group symmetries, we discard any rotations that yield identical rotated coordinates up to a small numerical tolerance. Overall  process is given in Figure \ref{fig:methodology_pipeline}.

\subsection{Structural, Cross-Scale, and Chemical Characteristics}

\paragraph{Elemental coverage and composition diversity.}
Across all structures, the dataset contains 41 unique chemical elements (highlighted in Figure \ref{fig:elementsAndMaterials}) spanning light elements, alkali and alkaline earth metals, transition metals, and heavy elements, reflecting broad chemical coverage. Elemental abundances are highly non-uniform, with hydrogen dominating the dataset at 55.07\% of all atoms, while the least abundant elements appear only sparsely (e.g., carbon at 0.05\%). The overall composition entropy is 2.177, corresponding to an effective element count of 8.82, indicating that although many elements are present, the effective chemical diversity is governed by a smaller subset that contributes most of the atomic mass. This imbalance reflects realistic materials statistics rather than artificially uniform sampling.

\paragraph{Element prevalence across atoms and structures.}
Elemental prevalence differs markedly when measured by atomic fraction versus structural presence. Hydrogen is both the most abundant element by atom count and the most prevalent across structures. In contrast, several heavier elements contribute meaningfully by atom count while appearing in relatively few structures, indicating concentrated inclusion within specific materials. Alkali metals such as Li, Na, and K rank among the most common elements both by atomic fraction and by number of structures, while transition metals such as V, Zn, and Cu appear less frequently but remain prominent contributors to atomic composition. This separation between atom-level dominance and structure-level coverage introduces meaningful heterogeneity, challenging models to reason consistently across both compositionally dense and sparse chemical regimes.

\begin{figure}
\centering 
\includegraphics[width=\columnwidth]{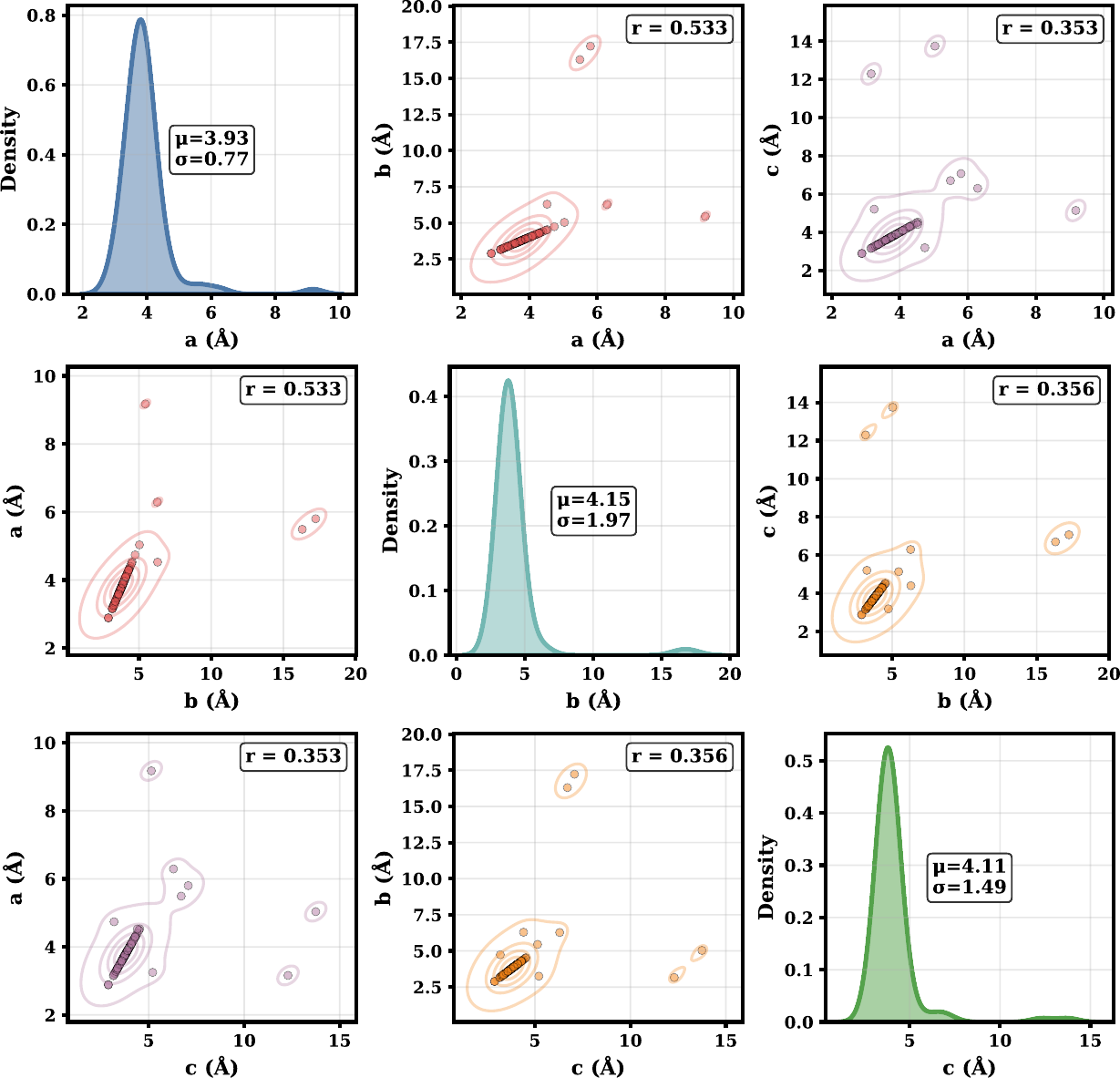} 
\caption{Distributions and pairwise relationships of lattice parameters ($a$, $b$, $c$) in the SCALAR dataset. Diagonal panels show kernel density estimates annotated with summary statistics, highlighting substantial variability and anisotropy across unit cells, with $b$ exhibiting the largest dispersion. Off-diagonal panels show pairwise scatter plots with Pearson correlation coefficients.}
\label{fig:datasetAnalysis} 
\end{figure}

\paragraph{Lattice parameter diversity.}
The diagonal panels of Figure~\ref{fig:datasetAnalysis} summarize the geometric variability of unit-cell dimensions through the distributions of lattice parameters $a$, $b$, and $c$ (in \AA). All three parameters exhibit broad support with comparable central tendencies, with means of $\mu_a=3.930$~\AA, $\mu_b=4.153$~\AA, and $\mu_c=4.112$~\AA, but markedly different dispersions. Parameter $a$ shows the narrowest distribution (standard deviation $0.768$~\AA, coefficient of variation $19.5\%$), while $b$ exhibits the greatest variability (standard deviation $1.970$~\AA, coefficient of variation $47.4\%$), and $c$ displays intermediate spread (standard deviation $1.489$~\AA, coefficient of variation $36.2\%$). The observed ranges ($a\in[2.884,9.174]$~\AA, $b\in[2.884,17.230]$~\AA, $c\in[2.884,13.747]$~\AA) indicate substantial diversity in unit-cell shape and anisotropy across the dataset, extending from compact, near-cubic cells to highly elongated geometries.

\paragraph{Coupling between lattice dimensions.}
The off-diagonal panels of Figure~\ref{fig:datasetAnalysis} characterize pairwise relationships between lattice parameters and reveal weak-to-moderate positive correlations. The strongest coupling is observed between $a$ and $b$ ($r=0.533$, $R^2=0.284$), with a linear trend indicating that increases in $a$ are typically accompanied by larger $b$ values. In contrast, correlations between $b$ and $c$ ($r=0.356$, $R^2=0.127$) and between $c$ and $a$ ($r=0.353$, $R^2=0.124$) are substantially weaker, reflecting greater independence between these dimensions. These asymmetric correlation patterns are consistent with crystallographic constraints that partially couple lattice parameters while still permitting significant heterogeneity in unit-cell geometry, underscoring the absence of a dominant length scale governing all three axes.

\paragraph{Growth of atom counts with radius.}
As structures are carved at increasing radii $R\in\{10,12,\dots,30\}$~\AA, atom counts increase strongly and approximately linearly with radius, exhibiting a high correlation ($r=0.967$, $R^2=0.935$). Mean atom counts rise from $377$ atoms at $R{=}10$~\AA\ to $10{,}152$ atoms at $R{=}30$~\AA, corresponding to an average increase of approximately $478$ atoms per additional angstrom of radius. Across the full dataset, atom counts span a wide dynamic range from $25$ to $18{,}123$ atoms, with a mean of $3{,}913$ atoms and substantial dispersion (coefficient of variation $89.7\%$). While the monotonic increase in mean atom count reflects controlled geometric growth, the large standard deviations and widening min--max ranges at larger radii indicate increasing structural variability across materials as system size grows.

\paragraph{Geometric growth of convex-hull volumes.}
Convex-hull volumes exhibit similarly strong scaling with radius, closely following a linear trend ($r=0.965$, $R^2=0.931$) with an average increase of approximately $5.09\times10^{3}$~\AA$^3$ per angstrom of radius. Mean hull volumes grow from $3{,}251$~\AA$^3$ at $R{=}10$~\AA\ to $107{,}352$~\AA$^3$ at $R{=}30$~\AA, spanning nearly two orders of magnitude across the dataset. Overall volumes range from $1{,}103$ to $109{,}893$~\AA$^3$, with a mean of $40{,}437$~\AA$^3$ and a coefficient of variation of $82.5\%$. The steady increase in both mean and spread with radius reflects systematic geometric expansion while accommodating material-dependent differences in shape, surface roughness, and atomic arrangement.

\begin{table*}[t]
\centering
\small
\setlength{\aboverulesep}{0.3pt}
\setlength{\belowrulesep}{0.3pt}
\setlength{\extrarowheight}{0.2pt}
\renewcommand{\arraystretch}{0.99}  
\caption{Forward CIF$\rightarrow$nanoparticle property prediction probing geometric scale generalization. Models must infer nanoparticle properties from crystal structure across increasing radii, isolating structural scale as a controlled distribution shift. Metrics capture hallucination (constraint and self-consistency violations), cross-scale consistency, physically grounded reasoning over scale trends, and numeric accuracy under 0/1/3-shot prompting. The results reveal systematic breakdowns in consistency and physical reasoning that are not apparent from accuracy alone, particularly under out-of-distribution scales. Shading highlights relative performance: \cellcolor{bestbg}green (optimal), \cellcolor{secondbg}blue (second-best), \cellcolor{poorbg}red (poorest) per column.}
\label{tab:scalar-main}
\resizebox{\linewidth}{!}{%
\begin{tabular}{lcccccccccc}
\toprule
 & Errors $\downarrow$ & \multicolumn{2}{c|}{Hallucination $\downarrow$} & \multicolumn{2}{c|}{Consistency $\downarrow$} & \multicolumn{2}{c|}{Reasoning $\uparrow$} & \multicolumn{2}{c|}{Accuracy $\downarrow$} & Parse Errors (\%) $\downarrow$ \\
\cmidrule(lr){2-2}\cmidrule(lr){3-4}\cmidrule(lr){5-6}\cmidrule(lr){7-8}\cmidrule(lr){9-10}\cmidrule(lr){11-11}
Model & All & ID & OOD & ID & OOD & ID & OOD & ID & OOD & All \\
\midrule
\multicolumn{11}{c}{\emph{0-shot}} \\
\midrule
Claude 3 Haiku & \cellcolor{poorbg}$0.755$ & \cellcolor{poorbg}$0.763$ & \cellcolor{poorbg}$0.747$ & \cellcolor{secondbg}$1.443$ & \cellcolor{bestbg}$1.617$ & $0.170$ & \cellcolor{poorbg}$0.177$ & $4.318$ & $4.511$ & \cellcolor{bestbg}$0.37\%$ \\
Seed 1.6 Flash & $0.711$ & $0.714$ & $0.716$ & $3.878$ & $4.068$ & $0.212$ & $0.219$ & $4.174$ & $4.367$ & \cellcolor{bestbg}$0.37\%$ \\
DeepSeek v3.2 & $0.628$ & $0.652$ & $0.647$ & $3.826$ & $4.036$ & $0.221$ & $0.228$ & $4.056$ & $4.253$ & \cellcolor{secondbg}$0.73\%$ \\
Gemini 3 Flash Preview & \cellcolor{secondbg}$0.147$ & \cellcolor{secondbg}$0.153$ & \cellcolor{bestbg}$0.161$ & $2.716$ & $2.906$ & $0.250$ & $0.256$ & $3.535$ & $3.725$ & $9.52\%$ \\
LLaMA 4 Maverick & $0.696$ & $0.722$ & $0.691$ & $2.700$ & $2.888$ & \cellcolor{poorbg}$0.167$ & $0.180$ & $4.252$ & $4.447$ & $32.97\%$ \\
Ministral 14B & $0.730$ & $0.738$ & $0.738$ & $2.653$ & $2.832$ & $0.176$ & $0.182$ & $4.249$ & $4.448$ & $7.69\%$ \\
Nemotron 3 Nano 30B & $0.463$ & $0.494$ & $0.493$ & \cellcolor{poorbg}$3.904$ & \cellcolor{poorbg}$4.095$ & $0.224$ & $0.200$ & \cellcolor{poorbg}$5.000$ & \cellcolor{poorbg}$5.000$ & \cellcolor{poorbg}$89.01\%$ \\
GPT-5 Mini & \cellcolor{bestbg}$0.107$ & \cellcolor{bestbg}$0.135$ & \cellcolor{secondbg}$0.166$ & \cellcolor{bestbg}$1.229$ & $2.759$ & \cellcolor{bestbg}$0.282$ & \cellcolor{bestbg}$0.279$ & \cellcolor{bestbg}$2.853$ & \cellcolor{bestbg}$3.085$ & $2.56\%$ \\
o3-mini & $0.375$ & $0.392$ & $0.406$ & $2.840$ & $3.032$ & $0.279$ & $0.277$ & $3.077$ & \cellcolor{secondbg}$3.195$ & $1.10\%$ \\
Grok 4.1 Fast & $0.188$ & $0.206$ & $0.230$ & $2.240$ & \cellcolor{secondbg}$2.627$ & \cellcolor{secondbg}$0.279$ & \cellcolor{secondbg}$0.277$ & \cellcolor{secondbg}$2.983$ & $3.197$ & $69.23\%$ \\
\midrule
\multicolumn{11}{c}{\emph{1-shot}} \\
\midrule
Claude 3 Haiku & $0.156$ & $0.426$ & $0.146$ & \cellcolor{secondbg}$1.360$ & \cellcolor{secondbg}$1.968$ & \cellcolor{poorbg}$0.106$ & \cellcolor{poorbg}$0.115$ & \cellcolor{poorbg}$4.531$ & $3.791$ & \cellcolor{bestbg}$0.00\%$ \\
Seed 1.6 Flash & \cellcolor{poorbg}$0.717$ & $0.721$ & \cellcolor{poorbg}$0.724$ & $3.661$ & $3.843$ & $0.196$ & $0.204$ & $4.213$ & $4.408$ & $1.47\%$ \\
DeepSeek v3.2 & $0.656$ & $0.666$ & $0.661$ & $3.887$ & $4.083$ & $0.218$ & $0.218$ & $4.098$ & $4.294$ & \cellcolor{secondbg}$0.37\%$ \\
Gemini 3 Flash Preview & \cellcolor{secondbg}$0.055$ & \cellcolor{secondbg}$0.058$ & \cellcolor{secondbg}$0.062$ & $2.219$ & $2.362$ & \cellcolor{secondbg}$0.280$ & $0.278$ & \cellcolor{bestbg}$2.773$ & \cellcolor{secondbg}$2.911$ & $8.06\%$ \\
LLaMA 4 Maverick & $0.436$ & $0.481$ & $0.434$ & $3.842$ & $4.014$ & $0.194$ & $0.202$ & $4.189$ & $4.392$ & $74.73\%$ \\
Ministral 14B & $0.714$ & \cellcolor{poorbg}$0.736$ & $0.708$ & $3.878$ & $4.060$ & $0.185$ & $0.190$ & $4.362$ & \cellcolor{poorbg}$4.541$ & $1.83\%$ \\
Nemotron 3 Nano 30B & $0.240$ & $0.266$ & $0.268$ & \cellcolor{poorbg}$4.153$ & \cellcolor{poorbg}$4.169$ & $0.239$ & $0.232$ & $4.009$ & $4.165$ & $83.15\%$ \\
GPT-5 Mini & $0.067$ & $0.081$ & $0.095$ & $2.540$ & $2.636$ & \cellcolor{bestbg}$0.281$ & \cellcolor{bestbg}$0.280$ & $2.818$ & \cellcolor{bestbg}$2.909$ & $1.47\%$ \\
o3-mini & $0.147$ & $0.155$ & $0.167$ & $3.187$ & $3.378$ & $0.280$ & \cellcolor{secondbg}$0.279$ & $3.195$ & $3.374$ & $0.73\%$ \\
Grok 4.1 Fast & \cellcolor{bestbg}$0.013$ & \cellcolor{bestbg}$0.013$ & \cellcolor{bestbg}$0.013$ & \cellcolor{bestbg}$0.685$ & \cellcolor{bestbg}$0.688$ & $0.278$ & $0.277$ & \cellcolor{secondbg}$2.801$ & $3.010$ & \cellcolor{poorbg}$94.14\%$ \\
\midrule
\multicolumn{11}{c}{\emph{3-shot}} \\
\midrule
Claude 3 Haiku & $0.291$ & $0.316$ & $0.277$ & $2.923$ & $3.129$ & $0.223$ & $0.209$ & $3.967$ & $4.077$ & \cellcolor{bestbg}$0.00\%$ \\
Seed 1.6 Flash & \cellcolor{poorbg}$0.672$ & \cellcolor{poorbg}$0.677$ & \cellcolor{poorbg}$0.679$ & $3.808$ & $3.988$ & $0.198$ & $0.209$ & $4.136$ & $4.333$ & \cellcolor{secondbg}$0.37\%$ \\
DeepSeek v3.2 & $0.426$ & $0.438$ & $0.444$ & $3.845$ & $4.062$ & $0.232$ & $0.239$ & $3.967$ & $4.171$ & $0.73\%$ \\
Gemini 3 Flash Preview & \cellcolor{bestbg}$0.044$ & \cellcolor{bestbg}$0.047$ & \cellcolor{bestbg}$0.046$ & \cellcolor{bestbg}$1.783$ & \cellcolor{bestbg}$1.952$ & $0.279$ & $0.277$ & \cellcolor{bestbg}$2.650$ & \cellcolor{bestbg}$2.789$ & $8.06\%$ \\
LLaMA 4 Maverick & $0.415$ & $0.448$ & $0.427$ & $3.621$ & $3.754$ & $0.201$ & $0.211$ & $4.061$ & $4.258$ & $79.49\%$ \\
Ministral 14B & $0.597$ & $0.622$ & $0.600$ & $3.923$ & \cellcolor{poorbg}$4.086$ & $0.199$ & $0.208$ & \cellcolor{poorbg}$4.234$ & \cellcolor{poorbg}$4.418$ & $1.10\%$ \\
Nemotron 3 Nano 30B & $0.288$ & $0.321$ & $0.297$ & \cellcolor{poorbg}$4.044$ & $4.005$ & \cellcolor{poorbg}$0.194$ & \cellcolor{poorbg}$0.182$ & $4.113$ & $4.246$ & $69.60\%$ \\
GPT-5 Mini & \cellcolor{secondbg}$0.052$ & \cellcolor{secondbg}$0.066$ & \cellcolor{secondbg}$0.074$ & \cellcolor{secondbg}$2.365$ & \cellcolor{secondbg}$2.566$ & \cellcolor{secondbg}$0.281$ & \cellcolor{secondbg}$0.279$ & \cellcolor{secondbg}$2.701$ & \cellcolor{secondbg}$2.842$ & $2.93\%$ \\
o3-mini & $0.160$ & $0.161$ & $0.181$ & $2.533$ & $2.818$ & $0.281$ & $0.279$ & $2.738$ & $2.957$ & $2.56\%$ \\
Grok 4.1 Fast & $0.182$ & $0.213$ & $0.146$ & -- & -- & \cellcolor{bestbg}$0.285$ & \cellcolor{bestbg}$0.286$ & $3.100$ & $3.320$ & \cellcolor{poorbg}$96.34\%$ \\
\bottomrule
\end{tabular}
}
\end{table*}

\section{Experiments}\label{sec:experiments}
We evaluate large language model reliability as materials foundation models under controlled geometric scale variation using the SCALAR benchmark. Our experiments examine how state-of-the-art LLMs predict nanoparticle properties across scale shifts, measuring hallucination rates, numeric error, consistency, and reasoning quality in in-distribution and out-of-distribution regimes, the effect of physics-grounded Chain-of-Thought (CoT) prompting, and robustness on inverse retrieval. We evaluate three tasks—direct CIF$\rightarrow$property prediction (Section~\ref{sec:task1}), a CoT variant (Section~\ref{sec:task2}), and inverse retrieval (Section~\ref{sec:task3})—across multiple foundation models. We quantify reliability across five dimensions: \emph{Hallucination rate} captures predictions violating physical constraints (e.g., negative densities, volumes exceeding bounds) or exhibiting self-consistency failures across repeated queries. \emph{Consistency} measures mean standard deviation of numeric predictions across $N=5$ independent queries per sample. \emph{Reasoning quality} computes Spearman rank correlation between predicted property changes across nanoparticle radii and ground-truth deltas, with penalty proportional to mean absolute delta error. \emph{Accuracy} is mean absolute error (MAE) between predicted and ground-truth values for density, volume, and nearest-neighbor distance. \emph{Parse error rate} reports the fraction failing to produce valid JSON structures. Complete experimental details available at the repository.

For inverse retrieval (Section~\ref{sec:task3}), we report \emph{top-1 accuracy} (fraction selecting the correct candidate from $K$ options), \emph{majority accuracy} (correct selection under majority voting), \emph{physical distance} (normalized $L_2$ distance $\|\mathbf{p}_{\text{target}} - \mathbf{p}_{\text{proposed}}\| / \|\mathbf{p}_{\text{target}}\|$ between target and proposed property vectors), and \emph{physical regret} (suboptimality as difference between proposed candidate's distance and minimum achievable). We apply log transformation $\log_{10}(x + 1)$ to metrics except parse errors (Sections~\ref{sec:task1}--\ref{sec:task2}), saturating values exceeding $10^5$; for delta metrics (Section~\ref{sec:task2}), we use signed variant $\text{sign}(x) \cdot \log_{10}(|x| + 1)$ to preserve directionality.

\begin{figure}[ht] 
\centering 
\includegraphics[width=\columnwidth]{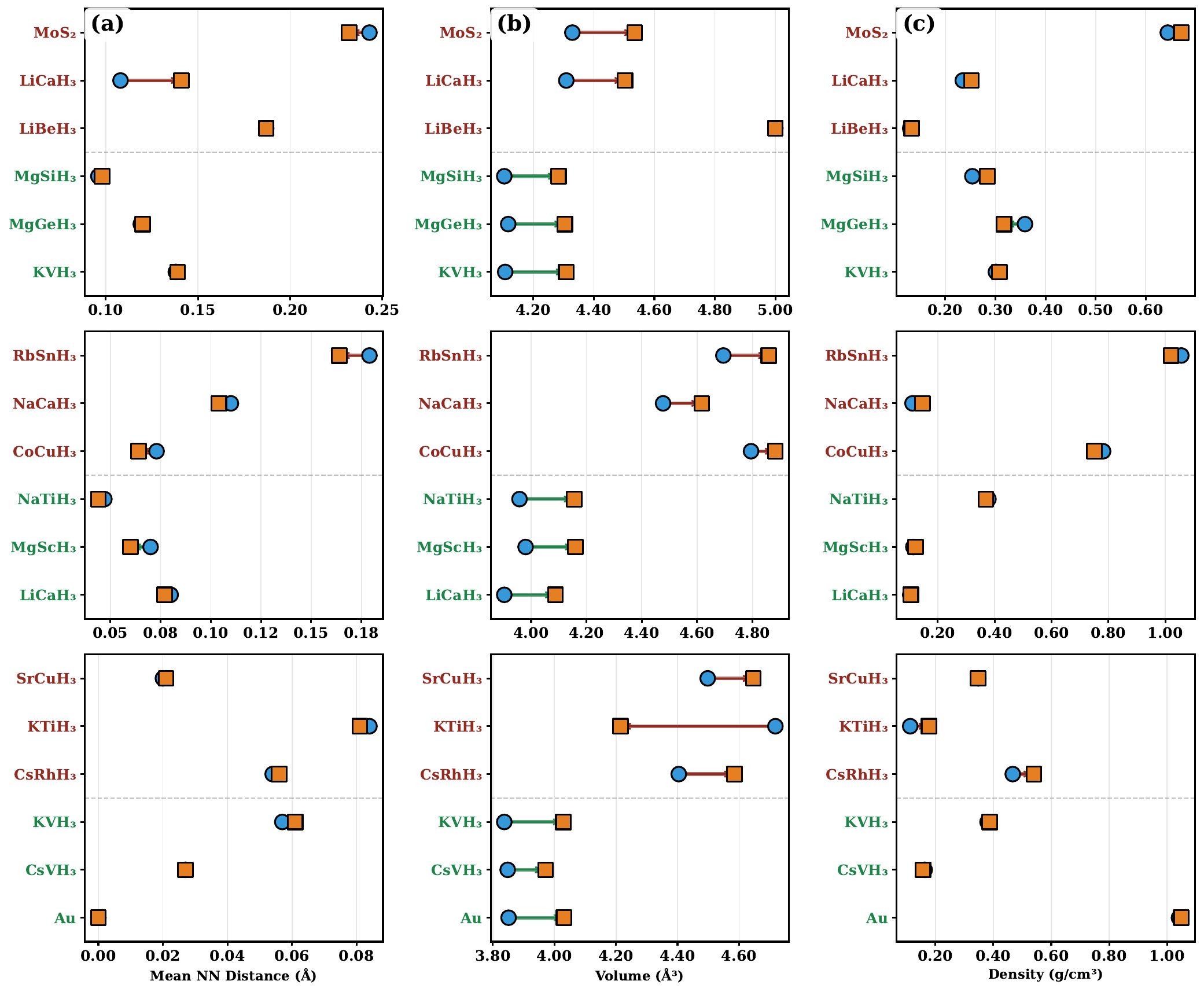} 
\caption{Material-level sensitivity analysis for geometric scale extrapolation. Rows correspond to 0-, 1-, 3-shot, and columns to mean nearest-neighbor distance, unit cell volume, density. For each regime, we show the top-3 best-performing (green) and worst-3 performing (red) materials. Circles (ID) and squares (OOD) are connected by lines whose length and direction indicate the extrapolation gap magnitude and sign.}
\label{fig:materialSensivity} 
\end{figure}

\subsection{Task 1: CIF$\rightarrow$Nanoparticle Property Prediction}
\label{sec:task1}
Within SCALAR, Table~\ref{tab:scalar-main} evaluates forward CIF$\rightarrow$nanoparticle property prediction as a controlled probe of geometric scale generalization by explicitly isolating structural scale as a distribution shift through in-distribution and out-of-distribution radii under 0/1/3-shot prompting. Across models, SCALAR reveals that hallucination rates and cross-scale inconsistency increase sharply under OOD scales, even when numeric accuracy degrades only modestly, showing that surface-level error metrics systematically understate failures in structural and physical reasoning. The reasoning metric, which captures alignment between predicted and gold scale trends, varies widely across models and is frequently decoupled from both accuracy and hallucination, indicating distinct failure modes in physical trend extrapolation that are most pronounced in the zero-shot regime. Few-shot prompting improves average performance for some models, but gains are highly model-dependent and non-monotonic, with some models exhibiting unstable behavior, degraded OOD consistency, or increased parse failures; critically, strong ID performance does not reliably translate to OOD robustness. Overall, SCALAR demonstrates that geometric scale shifts expose systematic breakdowns in physical reasoning and cross-scale consistency that are not captured by accuracy alone.

\textbf{Material-wise Error Heterogeneity Across Scale Regimes.}
Figure~\ref{fig:materialSensivity} reveals strong, property- and material-dependent sensitivity to geometric scale extrapolation that persists across prompting regimes. In the 0-shot setting, top-performing materials exhibit relatively small and stable ID--OOD gaps for mean nearest-neighbor distance (typically below 3\%), while several worst-performing materials show sharply amplified errors, including over 30\% relative increases for LiCaH$_3$ and sign reversals for MoS$_2$, indicating inconsistent geometric reasoning rather than uniform degradation. Volume errors grow more systematically with scale across both top and bottom materials, but density exhibits the largest heterogeneity, with some materials showing large non-monotonic shifts (e.g., MgGeH$_3$ and KTiH$_3$) even when distance and volume remain stable. Few-shot prompting reduces variance for some materials but does not eliminate these effects; under 3-shot prompting, extreme density extrapolation errors persist (e.g., $>50\%$ for KTiH$_3$), and sign-inconsistent trends remain common. These results demonstrate that scale-induced failures are strongly structure-dependent and property-specific, and that apparent robustness under aggregate metrics masks systematic extrapolation errors tied to material geometry. By exposing these fine-grained, material-level breakdowns under controlled scale transformations, SCALAR provides a principled lens for diagnosing geometric generalization failures that are invisible to model-averaged accuracy alone.

\begin{figure}
    \centering
    \includegraphics[width=1\linewidth]{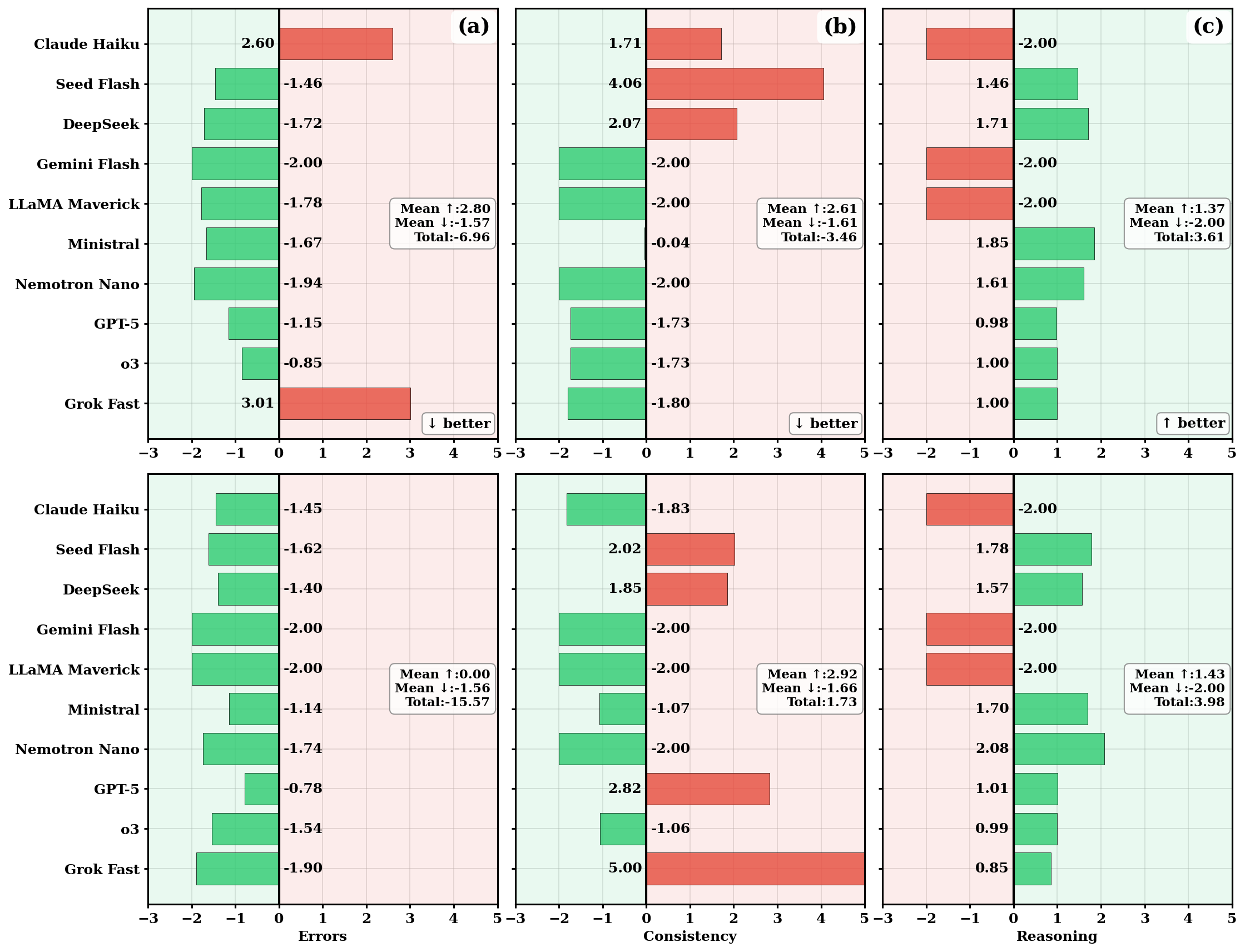}
    \caption{CoT prompting effect analysis. Horizontal diverging bar charts show the CoT impact (log-scaled $\Delta$) for ten models across three metrics under 1-shot and 3-shot regimes. Bars extend from a central zero line: leftward bars (negative $\Delta$, green) indicate that CoT improves the target metric, while rightward bars (positive $\Delta$, red) indicate that CoT worsens it. Metric direction differs by outcome: for \emph{Errors} and \emph{Consistency}, decreases are beneficial ($\downarrow$ better), whereas for \emph{Reasoning}, increases are beneficial ($\uparrow$ better).}
    \label{fig:butterflyCoT}
\end{figure}

\subsection{Task 2: Chain-of-Thought CIF$\rightarrow$Property Prediction}
\label{sec:task2}
Figure~\ref{fig:butterflyCoT} visualizes the effect of CoT prompting on cross-scale robustness using signed, log-scaled deltas, revealing highly non-uniform and metric-dependent behavior across models and shot regimes. In the 1-shot setting, CoT yields a net reduction in Errors and Consistency violations (negative total $\Delta$), driven by broad improvements across most models, but this comes at the cost of degraded reasoning for a non-trivial subset, with several models exhibiting maximal negative reasoning deltas despite improved error profiles. Conversely, reasoning shows a positive aggregate gain in both 1-shot and 3-shot regimes, yet these gains are frequently decoupled from improvements in Errors or Consistency, highlighting that better intermediate explanations do not reliably translate into more stable or physically consistent predictions. In the 3-shot setting, CoT uniformly reduces Errors across all models, but Consistency exhibits mixed behavior, with several models experiencing large degradations that offset gains elsewhere. Overall, the butterfly plots expose the trade-offs induced by CoT: while it can strengthen trend-level physical reasoning, it often introduces instability or inconsistency, underscoring that CoT is not a monotonic or universally beneficial intervention. By making these trade-offs explicit under controlled geometric scale shifts, SCALAR enables fine-grained diagnosis of when and how reasoning-style prompting helps or harms physical robustness beyond aggregate accuracy metrics.

\subsection{Task 3: Inverse Retrieval from Nanoparticle Properties}
\label{sec:task3}
Table~\ref{tab:inverseRetrieval} highlights substantial variation in inverse retrieval performance under geometric scale shifts. With $k=3$ candidates, Grok achieves the strongest performance (top-1 accuracy $0.808$ ID, $0.793$ OOD; physical regret $0.145$ ID, $0.209$ OOD), indicating correct selection and physically faithful predictions. Claude attains only $0.330$ top-1 accuracy with much larger physical distance ($0.742$ ID), suggesting failures reflect poor alignment with physical properties rather than classification errors alone. As task difficulty increases with $k=5$ candidates, top-1 accuracy drops across models (e.g., GPT-5 from $0.726\!\rightarrow\!0.622$ ID, $0.678\!\rightarrow\!0.573$ OOD) while physical regret increases, particularly under OOD scales. Notably, Nemotron maintains low physical distance at $k=5$ ($0.114$ ID, $0.176$ OOD) despite moderate accuracy, illustrating that physically consistent predictions do not always translate into correct retrieval. These trends demonstrate that SCALAR's inverse retrieval exposes scale-induced degradation in both discriminative accuracy and physical fidelity, revealing failure modes invisible to accuracy-only evaluation.

\begin{table}
\centering
\small
\caption{Inverse retrieval under geometric scale shift. Models are given nanoparticle properties and must identify the correct CIF among candidates, testing whether physical predictions are sufficiently informative and consistent to support discriminative structure matching. We report top-1 accuracy and majority-vote accuracy across repeats, along with physically grounded error measures: mean normalized L2 distance between predicted and target properties, and mean physical regret (distance to the selected candidate minus the best possible candidate).}
\label{tab:inverseRetrieval}
\resizebox{\linewidth}{!}{%
\begin{tabular}{lccccccccc}
\toprule
 & \multicolumn{2}{c|}{Top-1 Acc $\uparrow$} & \multicolumn{2}{c|}{Majority Acc $\uparrow$} & \multicolumn{2}{c|}{Phys Dist $\downarrow$} & \multicolumn{2}{c|}{Phys Regret $\downarrow$} & Parse Errors (\%) $\downarrow$ \\
\cmidrule(lr){2-3}\cmidrule(lr){4-5}\cmidrule(lr){6-7}\cmidrule(lr){8-9}\cmidrule(lr){10-10}
Model & ID & OOD & ID & OOD & ID & OOD & ID & OOD & All \\
\midrule
\multicolumn{10}{c}{\emph{3 Candidates}} \\
\midrule
Claude 3 Haiku & \cellcolor{poorbg}$0.330$ & \cellcolor{poorbg}$0.339$ & \cellcolor{poorbg}$0.330$ & \cellcolor{poorbg}$0.339$ & \cellcolor{poorbg}$0.742$ & \cellcolor{poorbg}$0.631$ & \cellcolor{poorbg}$0.742$ & \cellcolor{poorbg}$0.631$ & \cellcolor{bestbg}$0.11\%$ \\
Seed 1.6 Flash & $0.480$ & $0.485$ & $0.480$ & $0.485$ & $0.393$ & $0.413$ & $0.393$ & $0.413$ & $10.33\%$ \\
DeepSeek v3.2 & $0.678$ & $0.654$ & $0.678$ & $0.654$ & $0.271$ & $0.448$ & $0.271$ & $0.448$ & $29.34\%$ \\
Gemini 3 Flash Preview & $0.641$ & $0.609$ & $0.641$ & $0.609$ & $0.238$ & $0.285$ & $0.238$ & $0.285$ & \cellcolor{bestbg}$0.11\%$ \\
LLaMA 4 Maverick & $0.611$ & $0.605$ & $0.611$ & $0.605$ & $0.527$ & \cellcolor{bestbg}$0.093$ & $0.527$ & \cellcolor{bestbg}$0.093$ & \cellcolor{poorbg}$89.89\%$ \\
Ministral 14B & $0.456$ & $0.384$ & $0.456$ & $0.384$ & $0.659$ & $0.482$ & $0.659$ & $0.482$ & $74.18\%$ \\
Nemotron 3 Nano 30B & $0.723$ & $0.645$ & $0.723$ & $0.645$ & \cellcolor{bestbg}$0.144$ & \cellcolor{secondbg}$0.154$ & \cellcolor{bestbg}$0.144$ & \cellcolor{secondbg}$0.154$ & $1.54\%$ \\
GPT-5 Mini & \cellcolor{secondbg}$0.726$ & \cellcolor{secondbg}$0.678$ & \cellcolor{secondbg}$0.726$ & \cellcolor{secondbg}$0.678$ & $0.225$ & $0.280$ & $0.225$ & $0.280$ & $0.44\%$ \\
o3-mini & $0.672$ & $0.629$ & $0.672$ & $0.629$ & $0.193$ & $0.261$ & $0.193$ & $0.261$ & \cellcolor{secondbg}$0.33\%$ \\
Grok 4.1 Fast & \cellcolor{bestbg}$0.808$ & \cellcolor{bestbg}$0.793$ & \cellcolor{bestbg}$0.808$ & \cellcolor{bestbg}$0.793$ & \cellcolor{secondbg}$0.145$ & $0.209$ & \cellcolor{secondbg}$0.145$ & $0.209$ & $1.76\%$ \\
\midrule
\multicolumn{10}{c}{\emph{5 Candidates}} \\
\midrule
Claude Haiku & \cellcolor{poorbg}$0.235$ & \cellcolor{poorbg}$0.185$ & \cellcolor{poorbg}$0.235$ & \cellcolor{poorbg}$0.185$ & $0.431$ & $0.531$ & $0.431$ & $0.531$ & \cellcolor{bestbg}$0.00\%$ \\
Seed Flash & $0.356$ & $0.363$ & $0.356$ & $0.363$ & $0.592$ & $0.525$ & $0.592$ & $0.525$ & $5.60\%$ \\
DeepSeek& $0.508$ & $0.466$ & $0.508$ & $0.466$ & $0.443$ & $0.374$ & $0.443$ & $0.374$ & $29.56\%$ \\
Gemini Flash & $0.490$ & $0.466$ & $0.490$ & $0.466$ & $0.346$ & $0.392$ & $0.346$ & $0.392$ & \cellcolor{secondbg}$0.22\%$ \\
LLaMA Maverick & $0.389$ & $0.440$ & $0.389$ & $0.440$ & $0.292$ & $0.281$ & $0.292$ & $0.281$ & $77.58\%$ \\
Ministral& $0.256$ & $0.230$ & $0.256$ & $0.230$ & \cellcolor{poorbg}$0.810$ & \cellcolor{poorbg}$0.795$ & \cellcolor{poorbg}$0.810$ & \cellcolor{poorbg}$0.795$ & \cellcolor{poorbg}$84.29\%$ \\
Nemotron Nano& \cellcolor{secondbg}$0.624$ & \cellcolor{secondbg}$0.596$ & \cellcolor{secondbg}$0.624$ & \cellcolor{secondbg}$0.596$ & \cellcolor{bestbg}$0.114$ & \cellcolor{bestbg}$0.176$ & \cellcolor{bestbg}$0.114$ & \cellcolor{bestbg}$0.176$ & $6.37\%$ \\
GPT-5 & $0.622$ & $0.573$ & $0.622$ & $0.573$ & $0.290$ & $0.297$ & $0.290$ & $0.297$ & \cellcolor{secondbg}$0.22\%$ \\
o3 & $0.537$ & $0.490$ & $0.537$ & $0.490$ & $0.324$ & $0.421$ & $0.324$ & $0.421$ & $0.33\%$ \\
Grok Fast & \cellcolor{bestbg}$0.753$ & \cellcolor{bestbg}$0.664$ & \cellcolor{bestbg}$0.753$ & \cellcolor{bestbg}$0.664$ & \cellcolor{secondbg}$0.277$ & \cellcolor{secondbg}$0.240$ & \cellcolor{secondbg}$0.277$ & \cellcolor{secondbg}$0.240$ & $1.21\%$ \\
\bottomrule
\end{tabular}
}
\end{table}
\subsection{Geometry-Native and Physics-Based Baseline Comparisons}

To anchor task difficulty, we evaluate two complementary baselines. \textit{Physics-based analytical predictions} using unit cell CIFs with volume scaling laws ($N_{\text{atoms}} \propto R^3/V_{\text{cell}}$, density from bulk mass/volume) achieve low relative errors on geometric properties (1.9\% for nearest-neighbor distance, 14.7\% for volume) but larger errors on extensive properties (30.3\% for mass, 35.3\% for density). \textit{Geometry-native graph neural networks}—SchNet~\citep{schnet}, CGCNN~\citep{xie2018crystal}, and E(3)NN \citep{batzner20223}—trained on XYZ structures exhibit substantial scale-dependent degradation: E(3)NN's atom count MAPE increases from 106\% (ID) to 313\% (OOD), CGCNN degrades from 185\% to 464\%. These results demonstrate that \textit{geometric scale generalization challenges both text-based and geometry-aware paradigms}, establishing that (i) supervised models with geometric inductive biases do not trivially solve scale shifts, (ii) analytical baselines suffice only for intensive properties, and (iii) SCALAR's diagnostic value extends beyond LLM-specific evaluation.

\section{Limitations}\label{sec:limitations}
While SCALAR enables controlled evaluation of scale-dependent reasoning and hallucination in materials foundation models, several limitations remain: (i) the benchmark isolates structural scale through deterministic supercell expansion but does not capture disorder, defects, or thermal motion; (ii) evaluated properties are derived from classical geometric computations and do not reflect quantum-mechanical observables or downstream discovery performance; (iii) although target properties are rotation-invariant, ID/OOD rotation splits primarily probe text parsing robustness rather than geometric reasoning; (iv) the overlap between ``hallucination'' and ``consistency'' metrics may amplify correlated failure modes, parse error exclusion may bias results, and log-transformation masks absolute error magnitudes; (vi) the dataset includes a substantial fraction of hydrogen-containing materials relevant to energy storage and catalysis applications, and generalization of conclusions across hydride versus non-hydride chemistries requires further investigation and (vii) CoT prompting may introduce formatting instability that complicates attribution of improvements to genuine physical understanding.

\section{Conclusion}
We introduced SCALAR, a benchmark built from DFT-relaxed crystalline structures validated against experimental data for systematically probing geometric scale generalization, structure hallucination, and consistency in materials foundation models, through CIF$\rightarrow$nanoparticle property prediction, CoT reasoning, and inverse retrieval tasks. We show that many state-of-the-art models exhibit sharp, model-dependent breakdowns under scale variation even when aggregate accuracy appears strong, with hallucination tightly coupled to failures in cross-scale consistency and physically grounded reasoning, and with few-shot and CoT prompting yielding heterogeneous and non-guaranteed improvements. By framing hallucination as a consequence of geometric and chemical inconsistency under controlled structural transformations, and by introducing physically motivated distance and regret metrics that distinguish near-miss errors from implausible predictions, SCALAR complements traditional accuracy-based benchmarks and provides a principled setting for diagnosing and mitigating failures under realistic structural distribution shifts, paving the way toward more reliable deployment of foundation models in materials science.

\bibliography{main}

@incollection{tarantino2017structural,
  title={Structural modelling at the Micro-, Meso-, and Nanoscales},
  author={Tarantino, Angelo Marcello and Kaplunov, Julius and Luciano, Raimondo and Maiorana, Carmelo and Rousakis, Theodoros C and Willam, Kaspar and others},
  booktitle={Modelling and Simulation in Engineering},
  volume={2017},
  pages={1--3},
  year={2017},
  publisher={Hindawi}
}

@book{kittel2018introduction,
  title={Introduction to solid state physics},
  author={Kittel, Charles and McEuen, Paul},
  year={2018},
  publisher={John Wiley \& Sons}
}

@article{ashcroft1976solid,
  title={Solid state physics holt},
  author={Ashcroft, Neil W and Mermin, N David},
  journal={Rinehart and Winston, New York},
  volume={19761},
  pages={12},
  year={1976}
}

@article{pizzagalli2001structure,
  title={Structure and stability of germanium nanoparticles},
  author={Pizzagalli, Laurent and Galli, Giulia and Klepeis, John E and Gygi, Francois},
  journal={Physical Review B},
  volume={63},
  number={16},
  pages={165324},
  year={2001},
  publisher={APS}
}

@article{bera2010quantum,
  title={Quantum dots and their multimodal applications: a review},
  author={Bera, Debasis and Qian, Lei and Tseng, Teng-Kuan and Holloway, Paul H},
  journal={Materials},
  volume={3},
  number={4},
  pages={2260--2345},
  year={2010},
  publisher={Molecular Diversity Preservation International}
}

@article{vergara2017microed,
  title={MicroED structure of Au146 (p-MBA) 57 at subatomic resolution reveals a twinned FCC cluster},
  author={Vergara, Sandra and Lukes, Dylan A and Martynowycz, Michael W and Santiago, Ulises and Plascencia-Villa, German and Weiss, Simon C and de La Cruz, M Jason and Black, David M and Alvarez, Marcos M and Lopez-Lozano, Xochitl and others},
  journal={The Journal of Physical Chemistry Letters},
  volume={8},
  number={22},
  pages={5523--5530},
  year={2017},
  publisher={ACS Publications}
}

@article{li2023critical,
  title={A critical examination of robustness and generalizability of machine learning prediction of materials properties},
  author={Li, Kangming and DeCost, Brian and Choudhary, Kamal and Greenwood, Michael and Hattrick-Simpers, Jason},
  journal={npj Computational Materials},
  volume={9},
  number={1},
  pages={55},
  year={2023},
  publisher={Nature Publishing Group UK London}
}

@article{gleason2024random,
  title={Random forest prediction of crystal structure from electron diffraction patterns incorporating multiple scattering},
  author={Gleason, Samuel P and Rakowski, Alexander and Ribet, Stephanie M and Zeltmann, Steven E and Savitzky, Benjamin H and Henderson, Matthew and Ciston, Jim and Ophus, Colin},
  journal={Physical Review Materials},
  volume={8},
  number={9},
  pages={093802},
  year={2024},
  publisher={APS}
}

@article{anosova2024importance,
  title={The importance of definitions in crystallography},
  author={Anosova, Olga and Kurlin, Vitaliy and Senechal, Marjorie},
  journal={IUCrJ},
  volume={11},
  number={4},
  pages={453--463},
  year={2024},
  publisher={International Union of Crystallography}
}

@article{zhang2023surface,
  title={Surface reconstruction of CsPbBr3 nanocrystals by the ligand engineering approach for achieving high quantum yield and improved stability},
  author={Zhang, Yu and Hou, Guangning and Wu, Yong and Chen, Maosheng and Dai, Yannan and Liu, Shaohua and Zhao, Qingbiao and Lin, Hechun and Fang, Junfeng and Jing, Chengbin and others},
  journal={Langmuir},
  volume={39},
  number={17},
  pages={6222--6230},
  year={2023},
  publisher={ACS Publications}
}

@article{jain2013commentary,
  title={Commentary: The Materials Project: A materials genome approach to accelerating materials innovation},
  author={Jain, Anubhav and Ong, Shyue Ping and Hautier, Geoffroy and Chen, Wei and Richards, William Davidson and Dacek, Stephen and Cholia, Shreyas and Gunter, Dan and Skinner, David and Ceder, Gerbrand and others},
  journal={APL Materials},
  volume={1},
  number={1},
  year={2013},
  publisher={AIP Publishing}
}

@article{baig2021nanomaterials,
  title={Nanomaterials: A review of synthesis methods, properties, recent progress, and challenges},
  author={Baig, Nadeem and Kammakakam, Irshad and Falath, Wail},
  journal={Materials Advances},
  volume={2},
  number={6},
  pages={1821--1871},
  year={2021},
  publisher={Royal Society of Chemistry}
}

@article{ye2024strongly,
  title={Strongly-confined colloidal lead-halide perovskite quantum dots: from synthesis to applications},
  author={Ye, Junzhi and Gaur, Deepika and Mi, Chenjia and Chen, Zijian and Fern{\'a}ndez, Iago L{\'o}pez and Zhao, Haitao and Dong, Yitong and Polavarapu, Lakshminarayana and Hoye, Robert LZ},
  journal={Chemical Society Reviews},
  volume={53},
  number={16},
  pages={8095--8122},
  year={2024},
  publisher={Royal Society of Chemistry}
}

@article{kirklin2015open,
  title={The Open Quantum Materials Database (OQMD): assessing the accuracy of DFT formation energies},
  author={Kirklin, Scott and Saal, James E and Meredig, Bryce and Thompson, Alex and Doak, Jeff W and Aykol, Muratahan and R{\"u}hl, Stephan and Wolverton, Chris},
  journal={npj Computational Materials},
  volume={1},
  number={1},
  pages={1--15},
  year={2015},
  publisher={Nature Publishing Group}
}

@article{kim2025pubchem,
  title={PubChem 2025 update},
  author={Kim, Sunghwan and Chen, Jie and Cheng, Tiejun and Gindulyte, Asta and He, Jia and He, Siqian and Li, Qingliang and Shoemaker, Benjamin A and Thiessen, Paul A and Yu, Bo and others},
  journal={Nucleic Acids Research},
  volume={53},
  number={D1},
  pages={D1516--D1525},
  year={2025},
  publisher={Oxford University Press}
}

@article{dimenet,
  title={Directional message passing for molecular graphs},
  author={Gasteiger, Johannes and Gro{\ss}, Janek and G{\"u}nnemann, Stephan},
  journal={arXiv preprint arXiv:2003.03123},
  year={2020}
}

@article{schnet,
  title={Schnet--a deep learning architecture for molecules and materials},
  author={Sch{\"u}tt, Kristof T and Sauceda, Huziel E and Kindermans, P-J and Tkatchenko, Alexandre and M{\"u}ller, K-R},
  journal={The Journal of Chemical Physics},
  volume={148},
  number={24},
  year={2018},
  publisher={AIP Publishing}
}

@inproceedings{shoemake1985animating,
  title={Animating rotation with quaternion curves},
  author={Shoemake, Ken},
  booktitle={Proceedings of the 12th Annual Conference on Computer Graphics and Interactive Techniques},
  pages={245--254},
  year={1985}
}

@article{sriram2024flowllm,
  title={FlowLLM: Flow matching for material generation with large language models as base distributions},
  author={Sriram, Anuroop and Miller, Benjamin and Chen, Ricky TQ and Wood, Brandon},
  journal={Advances in Neural Information Processing Systems},
  volume={37},
  pages={46025--46046},
  year={2024}
}

@inproceedings{klipfel2023equivariant,
  title={Equivariant message passing neural network for crystal material discovery},
  author={Klipfel, Astrid and Bouraoui, Zied and Peltre, Olivier and Fregier, Ya{\"e}l and Harrati, Najwa and Sayede, Adlane},
  booktitle={Proceedings of the AAAI Conference on Artificial Intelligence},
  volume={37},
  pages={14304--14311},
  year={2023}
}

@inproceedings{polat2025stress,
    title = "Stress-Testing Multimodal Foundation Models for Crystallographic Reasoning",
    author = "Polat, Can  and
      Kurban, Hasan  and
      Serpedin, Erchin  and
      Kurban, Mustafa",
    year = "2025",
    booktitle = "Proceedings of the 3rd Workshop on Towards Knowledgeable Foundation Models (KnowFM)",
    publisher = "Association for Computational Linguistics",
    doi = "10.18653/v1/2025.knowllm-1.5",
    pages = "49--58",
    ISBN = "979-8-89176-283-1",
}

@article{polat2025beyond,
  title={Beyond Atomic Geometry Representations in Materials Science: A Human-in-the-Loop Multimodal Framework},
  author={Polat, Can and Serpedin, Erchin and Kurban, Mustafa and Kurban, Hasan},
  journal={arXiv preprint arXiv:2506.00302},
  year={2025}
}

@article{kurban2026multimodal,
  title={Multimodal explainable artificial intelligence-driven analysis of quantum size effects in copper nanoclusters for hydrogen storage},
  author={Kurban, Mustafa and Polat, Can and Serpedin, Erchin and Kurban, Hasan},
  journal={International Journal of Hydrogen Energy},
  volume={201},
  pages={152950},
  year={2026},
  publisher={Elsevier}
}

@article{ramos2025review,
  title={A review of large language models and autonomous agents in chemistry},
  author={Ramos, Mayk Caldas and Collison, Christopher J and White, Andrew D},
  journal={Chemical science},
  year={2025},
  publisher={Royal Society of Chemistry}
}

@article{jablonka202314,
  title={14 examples of how LLMs can transform materials science and chemistry: a reflection on a large language model hackathon},
  author={Jablonka, Kevin Maik and Ai, Qianxiang and Al-Feghali, Alexander and Badhwar, Shruti and Bocarsly, Joshua D and Bran, Andres M and Bringuier, Stefan and Brinson, L Catherine and Choudhary, Kamal and Circi, Defne and others},
  journal={Digital Discovery},
  volume={2},
  number={5},
  pages={1233--1250},
  year={2023},
  publisher={Royal Society of Chemistry}
}

@article{mirza2025framework,
  title={A framework for evaluating the chemical knowledge and reasoning abilities of large language models against the expertise of chemists},
  author={Mirza, Adrian and Alampara, Nawaf and Kunchapu, Sreekanth and R{\'\i}os-Garc{\'\i}a, Marti{\~n}o and Emoekabu, Benedict and Krishnan, Aswanth and Gupta, Tanya and Schilling-Wilhelmi, Mara and Okereke, Macjonathan and Aneesh, Anagha and others},
  journal={Nature Chemistry},
  pages={1--8},
  year={2025},
  publisher={Nature Publishing Group UK London}
}

@article{m2024augmenting,
  title={Augmenting large language models with chemistry tools},
  author={M. Bran, Andres and Cox, Sam and Schilter, Oliver and Baldassari, Carlo and White, Andrew D and Schwaller, Philippe},
  journal={Nature Machine Intelligence},
  volume={6},
  number={5},
  pages={525--535},
  year={2024},
  publisher={Nature Publishing Group UK London}
}

@article{jablonka2024leveraging,
  title={Leveraging large language models for predictive chemistry},
  author={Jablonka, Kevin Maik and Schwaller, Philippe and Ortega-Guerrero, Andres and Smit, Berend},
  journal={Nature Machine Intelligence},
  volume={6},
  number={2},
  pages={161--169},
  year={2024},
  publisher={Nature Publishing Group UK London}
}

@article{zheng2023chatgpt,
  title={ChatGPT in scientific writing: a cautionary tale},
  author={Zheng, Haoyi and Zhan, Huichun},
  journal={The American Journal of Medicine},
  volume={136},
  number={8},
  pages={725--726},
  year={2023},
  publisher={Elsevier}
}

@article{ji2023survey,
  title={Survey of hallucination in natural language generation},
  author={Ji, Ziwei and Lee, Nayeon and Frieske, Rita and Yu, Tiezheng and Su, Dan and Xu, Yan and Ishii, Etsuko and Bang, Ye Jin and Madotto, Andrea and Fung, Pascale},
  journal={ACM computing surveys},
  volume={55},
  number={12},
  pages={1--38},
  year={2023},
  publisher={ACM New York, NY}
}

@article{maynez2020faithfulness,
  title={On faithfulness and factuality in abstractive summarization},
  author={Maynez, Joshua and Narayan, Shashi and Bohnet, Bernd and McDonald, Ryan},
  journal={arXiv preprint arXiv:2005.00661},
  year={2020}
}

@article{mishra2024foundational,
  title={Foundational large language models for materials research},
  author={Mishra, Vaibhav and Singh, Somaditya and Ahlawat, Dhruv and Zaki, Mohd and Bihani, Vaibhav and Grover, Hargun Singh and Mishra, Biswajit and Miret, Santiago and Krishnan, NM and others},
  journal={arXiv preprint arXiv:2412.09560},
  year={2024}
}

@article{xie2024darwin,
  title={Darwin 1.5: Large language models as materials science adapted learners},
  author={Xie, Tong and Wan, Yuwei and Liu, Yixuan and Zeng, Yuchen and Wang, Shaozhou and Zhang, Wenjie and Grazian, Clara and Kit, Chunyu and Ouyang, Wanli and Zhou, Dongzhan and others},
  journal={arXiv preprint arXiv:2412.11970},
  year={2024}
}

@article{moro2025multimodal,
  title={Multimodal foundation models for material property prediction and discovery},
  author={Moro, Viggo and Loh, Charlotte and Dangovski, Rumen and Ghorashi, Ali and Ma, Andrew and Chen, Zhuo and Kim, Samuel and Lu, Peter Y and Christensen, Thomas and Solja{\v{c}}i{\'c}, Marin},
  journal={Newton},
  volume={1},
  number={1},
  year={2025},
  publisher={Elsevier}
}

@article{kong2025mattertune,
  title={Mattertune: An integrated, user-friendly platform for fine-tuning atomistic foundation models to accelerate materials simulation and discovery},
  author={Kong, Lingyu and Shoghi, Nima and Hu, Guoxiang and Li, Pan and Fung, Victor},
  journal={arXiv preprint arXiv:2504.10655},
  year={2025}
}

@article{bickelhaupt2000kohn,
  title={Kohn-Sham density functional theory: predicting and understanding chemistry},
  author={Bickelhaupt, F Matthias and Baerends, Evert Jan},
  journal={Reviews in Computational Chemistry},
  pages={1--86},
  year={2000},
  publisher={Wiley Online Library}
}

@article{xie2018crystal,
  title={Crystal graph convolutional neural networks for an accurate and interpretable prediction of material properties},
  author={Xie, Tian and Grossman, Jeffrey C},
  journal={Physical review letters},
  volume={120},
  number={14},
  pages={145301},
  year={2018},
  publisher={APS}
}

@article{batzner20223,
  title={E (3)-equivariant graph neural networks for data-efficient and accurate interatomic potentials},
  author={Batzner, Simon and Musaelian, Albert and Sun, Lixin and Geiger, Mario and Mailoa, Jonathan P and Kornbluth, Mordechai and Molinari, Nicola and Smidt, Tess E and Kozinsky, Boris},
  journal={Nature communications},
  volume={13},
  number={1},
  pages={2453},
  year={2022},
  publisher={Nature Publishing Group UK London}
}
\bibliographystyle{icml2026}





\end{document}